\documentclass{article}

\usepackage{bm}
\usepackage{array,url,amssymb,bbding,cite}
\usepackage{spconf,amsmath,graphicx,setspace,epstopdf,booktabs}
\usepackage{tabularx, makecell, multirow, multicol, arydshln, stfloats}

\title{CIF-BASED COLLABORATIVE DECODING FOR END-TO-END CONTEXTUAL SPEECH RECOGNITION}

\name{Minglun Han$^{1,2,*}$, Linhao Dong$^{1,*}$, Shiyu Zhou$^{1}$, Bo Xu$^{1,2}$\thanks{This work is supported by the National Key Research and Development Program of China under No.2017YFB1002102.}\thanks{* denotes equal contribution to this work.}}
\address{
$^{1}$Institute of Automation, Chinese Academy of Sciences, Beijing, China\\
$^{2}$School of Artificial Intelligence, University of Chinese Academy of Sciences, Beijing, China\\
\small \tt \{hanminglun2018, donglinhao2015, zhoushiyu2013, xubo\}@ia.ac.cn}

\begin{document}
\ninept
\renewcommand{\baselinestretch}{0.8812} \normalsize

\maketitle

\begin{abstract}
End-to-end (E2E) models have achieved promising results on multiple speech recognition benchmarks, and shown the potential to become the mainstream. However, the unified structure and the E2E training hamper injecting context information into them for contextual biasing. Though contextual LAS (CLAS) gives an excellent all-neural solution, the degree of biasing to given contextual information is not explicitly controllable. In this paper, we focus on incorporating contextual information into the continuous integrate-and-fire (CIF) based model that supports contextual biasing in a more controllable fashion. Specifically, an extra context processing network is introduced to extract contextual embeddings, integrate acoustically relevant contextual information and decode the contextual output distribution, thus forming a collaborative decoding with the decoder of the CIF-based model. Evaluated on the named entity rich evaluation sets of HKUST/AISHELL-2, our method brings relative character error rate (CER) reduction of 8.83\%/21.13\% and relative named entity character error rate (NE-CER) reduction of 40.14\%/51.50\% when compared with a strong baseline. Besides, it keeps the performance on original evaluation set without degradation.
\end{abstract}

\begin{keywords}
End-to-end, contextual biasing, continuous integrate-and-fire, collaborative decoding
\end{keywords}

\section{Introduction}
\label{sec:introduction}

Benefiting from the joint training of all components, end-to-end models \cite{graves2012sequence,hannun2014deep,ChorowskiBSCB15,chan2016listen,ChiuSWPNCKWRGJL18,dong2018speech,lu2020exploring} have achieved competitive results on different scales of automatic speech recognition (ASR) datasets. However, since E2E models integrate all modules into a unified structure, the model flexibility of them degrades compared to the conventional systems. How to bias the recognition process towards contextual information \cite{scheiner2016voice} to enhance the capability of customization becomes a popular research field for current E2E models.

A number of representative approaches have emerged in injecting contextual information into E2E models in recent years. Shallow fusion \cite{williams2018contextual,zhao2019shallow,chen2019end} is one kind of methods which bias the decoding process of E2E models towards a independently built language model (LM) via adjusting the decoding scores in beam search. Similar to the shallow fusion based methods, on-the-fly (OTF) rescoring \cite{scheiner2016voice,hall2015composition,aleksic2015bringing} is also applied to E2E models by fusing a FST compiled with contextual n-grams into the search process \cite{pundak2018deep}. In contrast with the above methods whose biasing model is built independently from the E2E models, contextual LAS (CLAS) \cite{pundak2018deep} and its extensions \cite{bruguier2019phoebe,chen2019joint,alon2019contextual} fold an extra contextual module and the ASR module in an all-neural network to incorporate contextual information. Due to the strong modeling expressiveness of neural networks, CLAS achieves significant improvement in comparison with the OTF rescoring \cite{pundak2018deep}. However, the unified structure makes it unfeasible to explicitly control the degree of biasing during inference. Apart from above label-synchronous models, the all-neural contextual biasing method applied on RNN Transducer (RNN-T) \cite{jain2020contextual} also faces the similar situation. Thus, we aim to explore a more controllable contextual biasing method by decoupling neural contextual components from the unified structure. Besides, most previous all-neural contextual biasing approaches are experimented on English datasets with thousands of hours of speech. We wonder whether such methods can achieve comparable improvements on the Mandarin ASR datasets and on the datasets with less training data (e.g. hundreds of hours).

In this work, we focus on incorporating contextual information into a frame-synchronous E2E model -- Continuous Integrate-and-Fire (CIF) based model \cite{cif}, which uses a soft and monotonic alignment mechanism and supports the extraction of acoustic embeddings (each of them corresponds to an output unit). The extracted acoustic embedding $c_i$ provides a bridge for the CIF-based model to interact with contextual information to obtain the acoustically relevant contextual embedding $m_i$. Since $m_i$ could be consumed by a contextual decoder to get the contextual output distribution, the contextual decoder together with the original decoder of the CIF-based model performs a collaborative decoding with a tunable weight to get the final hypothesis. Differing from CLAS and its extensions, the tunable weight during inference endows our model with the ability to explicitly control the effect of contextual biasing. Differing from the OTF rescoring and shallow fusion, the neural network based structure and the direct interaction between the context and the extracted acoustic embedding enable our model to conduct an effective joint training.

To further improve our method, we propose two auxiliary techniques: 1) a gradient accumulation based training strategy to enable our model to see more diverse contextual information; 2) an attention scaling interpolation (ASI) technique to alleviate the over-biasing phenomenon which means the contextual information mistakenly biases irrelevant locations. Evaluated on named entity rich evaluation sets of two Mandarin ASR datasets which covers 167 and 1000 hours of training data, our method achieves relative NE-CER reduction ranging from 40.14\% to 51.50\% and relative CER reduction ranging from 8.83\% to 21.13\%. Meanwhile, our method does not degrade the performance on original evaluation set.

\section{Proposed Methods}
\label{sec:proposed methods}

\subsection{Continuous Integrate-and-Fire}
\label{ssec:continuous integrate-and-fire}

As shown in the dashed box in Fig.\ref{Fig.1}, continuous integrate-and-fire (CIF) is a middleware connecting the encoder and the decoder. The encoder receives the feature $X=\{x_1, x_2, ..., x_t, ..., x_T\}$ and produces the encoded output $H=\{h_1, h_2, ..., h_u, ..., h_U\}$. At each encoder step $u$, CIF accumulates the weight $\alpha_{u}$ derived from $h_u$ in a left-to-right manner, thus it could support online ASR. In the process of accumulation, the accumulated weight reaching the threshold (1.0) is considered as a sign of firing, and the encoder step $u$ corresponding to this firing is marked as an acoustic boundary step. The weight of the acoustic boundary step is divided into two parts: one for the integration of current label, and the other one for the integration of next label. Then, CIF integrates the relevant encoded outputs in the form of weighted-sum to obtain the acoustic embedding $c_i$ corresponding to current output label $y_i$. After the integration, the integrated acoustic embedding $c_i$ is fired to the decoder for prediction. With $c_i$ and last predicted label $y_{i-1}$, the decoder could predict the probability of current output label $y_i$:
\begin{equation}
    P(y_i|c_{\le{i}},y_{\le{i-1}}) = \mathrm{Decoder}(y_{i-1}, c_i)
\end{equation}

\subsection{Context Processing Network}
\label{ssec:CIF-based collaborative decoding}

\begin{figure*}[ht]
  \centering
  \includegraphics[scale=0.90]{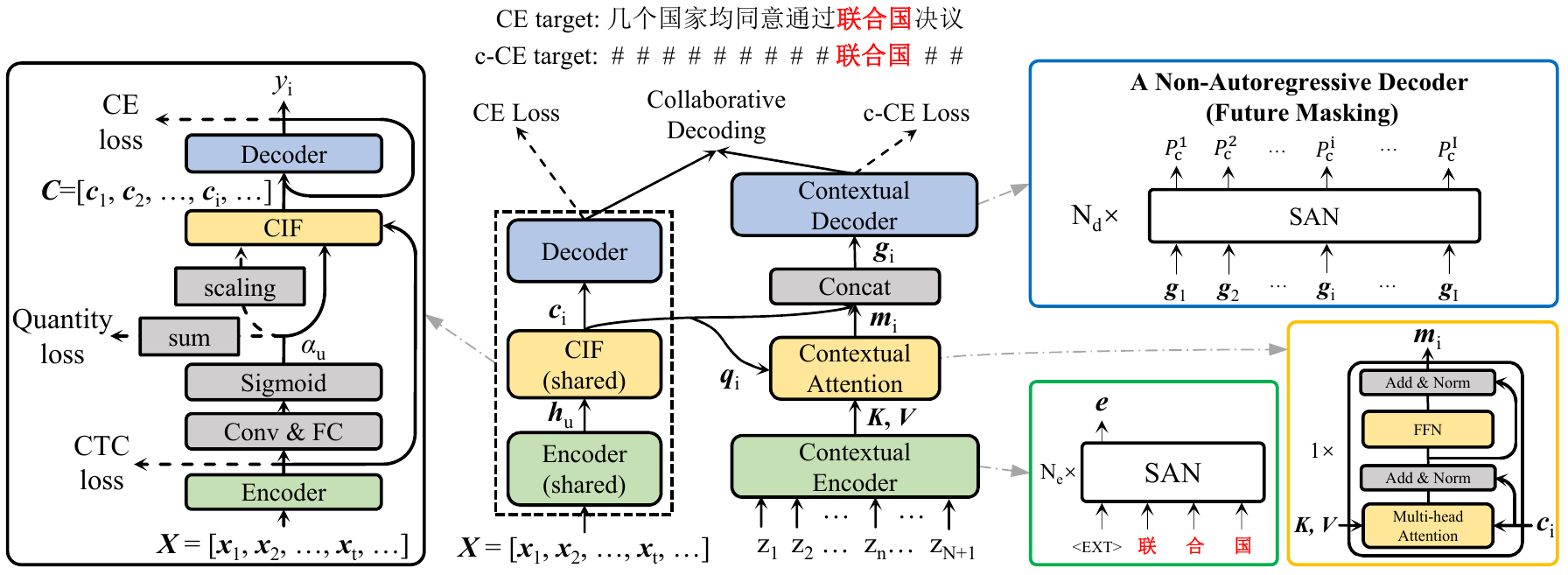}
  \vspace{-4.5pt}
  \caption{The collaborative decoding (ColDec) method on the CIF-based model: 1) the structure in the left solid box is the original CIF-based model; 2) the diagram in the middle shows the whole framework; 3) three solid boxes on the right side shows details of context processing network. A Chinese named entity meaning `the United Nations' in English is used as an example to help understand how contextual embeddings are extracted and how the target of c-CE loss is generated from the target of CE loss.}
  \label{Fig.1}
  \vspace{-13.5pt}
\end{figure*}

As depicted in Fig.\ref{Fig.1}, we introduce a context processing network to the original CIF-based model, and it is composed of three parts: the contextual encoder (c-encoder), the contextual attention (c-attention) and the contextual decoder (c-decoder).

Since the contextual information is commonly considered as the contextual segments (n-grams, queries and entities) \cite{scheiner2016voice,hall2015composition,pundak2018deep,VelikovichWSAMR18,huang2020class} in contextual speech recognition, the c-encoder is mainly used to extract the embeddings of these segments. Given a list of $N$ contextual segments ${Z=\{z_1, z_2, ..., z_N\}}$, the c-encoder encodes each segment to a vector. Here, we introduce two extra marks $<$EXT$>$ and $<$NO-BIAS$>$: $<$EXT$>$ is added to the beginning of each segment as a placeholder for extraction (as shown in the right part of Fig.\ref{Fig.1}), and `$<$EXT$><$NO-BIAS$>$' stands for an extra \textit{no-bias} option denoting no bias is needed. Thus, $N + 1$ embeddings $\{{e}_1, {e}_2, ..., {e}_{N+1}\}$ could be extracted from the $<$EXT$>$ output location of the c-encoder. With the $N+1$ embeddings obtained, the c-attention uses them as keys ($K$) and values ($V$), and uses the acoustic embedding $c_i$ from CIF as a query ($q_i$) for multi-head attention to produce current acoustically relevant contextual emdedding $m_i$. $c_i$ and $m_i$ are then concatenated as input $g_i$ to the c-decoder. The c-decoder is a non-autoregressive decoder \cite{cif} with future masking (each decoding step only attends to itself and its previous steps), whose output models the contextual output distribution $P_{c}{(y_i|g_{\le{i}})}$. The calculation process of $P_{c}$ is presented as follows (residual connections and layer normalization are omitted here):
\begin{align}
    K&=V=[ {e}_1, {e}_2, ..., {e}_{n}, ..., {e}_{N+1} ]^T\\
    Q&=[ c_1, c_2, ..., c_i, ..., c_I ]^T
\end{align}
\begin{align}
    f_i&=\mathrm{MultiHeadAttention}(c_i^T,K,V)\\
    m_i&=\mathrm{FeedForwardNetwork}(f_i)\\
    g_i&=[ c_i; \ m_i ]\\
    P_c(y_i|g_{\le{i}})&=\mathrm{NonAutoregressiveDecoder}(g_{i})
\end{align}

\subsection{Collaborative Decoding}
\label{sec:collaborative decoding}

Since the context processing network is driven by the incoming acoustic embedding $c_i$, the original decoder and the c-decoder predict their respective $i$-th output distributions synchronously, thus we call our method collaborative decoding (ColDec) in this work. 

In the training stage, the whole structure supports not only to be trained from scratch, but also to be trained by freezing a trained original CIF-based model and just updating the context processing network. In this work, we use the second way for faster training. In detail, the context processing network is trained with the contextual cross entropy (c-CE) loss, which is designed to enable the context processing network to interact with the acoustic embedding $c_i$ and extract the acoustically relevant contextual embedding $m_i$. The training target of c-CE loss is generated from original reference by keeping the characters in the overlapped part between reference and the given contextual segments unchanged, and masking the others with $<$NON$>$ (denoted as `\#') as shown at the top of Fig.1. For each training batch of the context processing network, we first use a tokenizer\footnote{\url{https://github.com/fxsjy/jieba}} to segment the reference of each utterance into some words (which is necessary for Chinese text), then we randomly sample a value $n$ from $\{1, 2, 3, 4\}$ and randomly extract an $n$-gram (using the sampled $n$) from each segmented reference. Finally, we randomly take 50\% of extracted n-grams as a list of contextual segments (including \textit{no-bias} option, denoted as c-batch in the later part).

During inference, the original decoder of the CIF-based model and the c-decoder conduct collaborative decoding using beam search as follows:
\begin{equation}
Y^{*} = \mathop{\arg\max}_{Y} \Bigr\{ {\sum\nolimits_{i}} \bigr( \mathrm{log}P(y_i|c_{\le{i}},y_{\le{i-1}}) + \lambda \mathrm{log}P_c(y_i|g_{\le{i}}) \bigr) \Bigr\}
\end{equation}
where $P(y_i|c_{\le{i}},y_{\le{i-1}})$ and $P_c(y_i|g_{\le{i}})$ represent the probabilities from the original decoder of the CIF-based model and c-decoder, respectively, and $\lambda$ is a tunable weight that controls the influence of contextual information.

\subsection{Auxiliary Techniques}
\label{sec:auxiliary techniques}

To conduct more effective learning for the context processing network, we apply a gradient accumulation based training strategy. For each training batch, our framework first generates multiple c-batches. Then, the c-CE losses corresponding to these c-batches are calculated independently, and their gradients are averaged for model updating. This strategy reuses the extracted acoustic embeddings from CIF, and enables our model to see more diverse context.

In the inference, we propose an attention scaling interpolation (ASI) method for collaborative decoding. This technique is used to relieve the over-biasing phenomenon (the contextual information mistakenly biases irrelevant locations). Since the interpolation weight $\lambda$ directly influences the contribution of contextual information during inference, we seek to weaken the bias of the decoding steps whose bias confidence is not strong enough. Here, we refer $1 - \alpha_{nb}^{i}$ ($\alpha_{nb}^{i}$ is the attention weight of \textit{no-bias} at step $i$) as the bias confidence. With $1 - \alpha_{nb}^{i}$ as the scaling factor of $\lambda$, the influence of steps with low bias confidence is suppressed, while the influence of steps with high bias confidence is almost kept. New interpolation weight becomes  $(1 - \alpha_{nb}^{i}) * \lambda$. Note that the attention weights of each step are averaged across all attention heads of c-attention:
\begin{equation}
{\alpha}^{i} = \sum_{m=1}^{M} {\alpha}_{m}^{i}
\end{equation}
where $M$ represents the number of attention heads in c-attention.

\section{Experiments}
\label{sec:experiments}

\subsection{Datasets and Metrics}
\label{sssec:datasets and metrics}

We experiment on two Mandarin datasets: HKUST \cite{liu2006hkust} and AISHELL-2 \cite{du2018aishell}. The data for training and development is the same with \cite{cif}. Apart from regular (original) evaluation sets called HKUST-RE and AS2-ios-RE (for \textit{test-ios} of AISHELL-2) here, we build named entity (NE) rich evaluation sets called HKUST-NE and AS2-ios-NE for the evaluation of contextual biasing. These NE rich evaluation sets are built as follows. First, HanLP\footnote{\url{https://github.com/hankcs/HanLP}} is used to detect NEs from all references in regular evaluation set. Second, the utterances that have NEs detected constitute NE rich evaluation set. The NE rich evaluation set and its corresponding NEs mimic an scenario where NEs are known as contextual information prior to recognition. The statistics about the NE rich evaluation sets is listed in Table \ref{tab:evaluation sets}, in which OOV rate refers to the proportion of out-of-vocabulary NEs (entities that are not present in the training set) in all NEs.

\begin{table}[h]
\vspace{-13.5pt}
\centering
\caption{Details of NE rich evaluation sets for both datasets.}
\vspace{4.5pt}
\renewcommand\arraystretch{0.83}
\newcommand{\tabincell}[2]{\begin{tabular}{@{}#1@{}}#2\end{tabular}}
\begin{tabular}{p{1.6cm}p{1.4cm}<{\centering}p{1.8cm}<{\centering}p{1.8cm}<{\centering}}
\toprule
{\tabincell{c}{\textbf{Evaluation}\\\textbf{Set}}} & {\tabincell{c}{\textbf{Number of}\\\textbf{Utterances}}} & {\tabincell{c}{\textbf{Number of}\\\textbf{Named Entities}}} & \tabincell{c}{\textbf{OOV} \textbf{Rate(\%)}} \\
\midrule
HKUST-NE & 726 & 280 & 32.14\\
AS2-ios-NE & 1206 & 691 & 35.60\\
\bottomrule
\end{tabular}
\label{tab:evaluation sets}
\vspace{-6pt}
\end{table}

We use character error rate (CER) and named entity character error rate (NE-CER) as the metric of the quality of general ASR and NE recognition. NE-CER is similar to NE-WER defined in \cite{garofolo19991998}, but use Chinese character rather than word as the basic unit for metric. The recognizer hypothesis and the reference are aligned first. Then, the calculation of NE-CER is implemented on all annotated NEs in references and their corresponding aligned segments in hypotheses by calculating the CER between them.

\subsection{Experimental Setup}
\label{ssec:experimental setup}

Input features are extracted with the same setup in \cite{dong2019self}. Speed perturbation \cite{ko2015audio} is applied to all training sets. Specaugment \cite{park2019specaugment} with the same setup in \cite{cif} are applied to all training experiments with big model setup. In our work, we generate 5230 output labels for AISHELL-2 and 3674 output labels for HKUST as \cite{cif}, and introduce three tokens for ColDec: 1) $<$EXT$>$ as the placeholder for extracting the vector of contextual segment; 2) $<$NO-BIAS$>$ as a sign of \textit{no-bias} option; 3) $<$NON$>$ as a mark denoting no label output.

We refer the models comprised of self-attention layers with $h=4$, $d_{model}=512$ and $d_{ff}=2048$ as the base model, and the models with $h=4$, $h_{model}=640$ and $d_{ff}=2560$ as the big model. For the CIF-based model on HKUST, we experiment with the same big setup in \cite{cif} and base setup. For the CIF-based model on AISHELL-2, we only experiment with the same big setup in \cite{cif}. As for context processing network, we use $N_e=2$ self-attention layers as the c-encoder, one layer with multi-head attention as the c-attention and $N_d=2$ self-attention layers as the c-decoder. The context processing network uses the same base or big setup for self-attention layers with the CIF-based model. During training, we apply residual dropout and attention dropout with $P_{drop}=0.1$ for self-attention layers, and use uniform label smoothing in \cite{chorowski2016towards} with $\beta=0.1$ for all experiments. After training, the average of the latest 10 saved models is taken as the final model for evaluation. During inference, we first generate hypotheses using beam search with beam size $10$, and then perform rescoring with a language model trained with the text of training set \cite{cif}. The language model interpolation weight in LM rescoring is set to $0.2$ for all datasets.

\section{Results}
\label{sec:results}

\subsection{Results on HKUST}
\label{ssec:results on hkust}

In this section, we first investigate the effect brought by the two auxiliary techniques on our CIF-based collaborative decoding (ColDec) method. Then, we apply the investigated best ColDec setup on the CIF-based model to obtain the performance improvements on the big model.

\begin{figure}[!htb]
  \vspace{-4.5pt}
  \centering
  \includegraphics[scale=0.35]{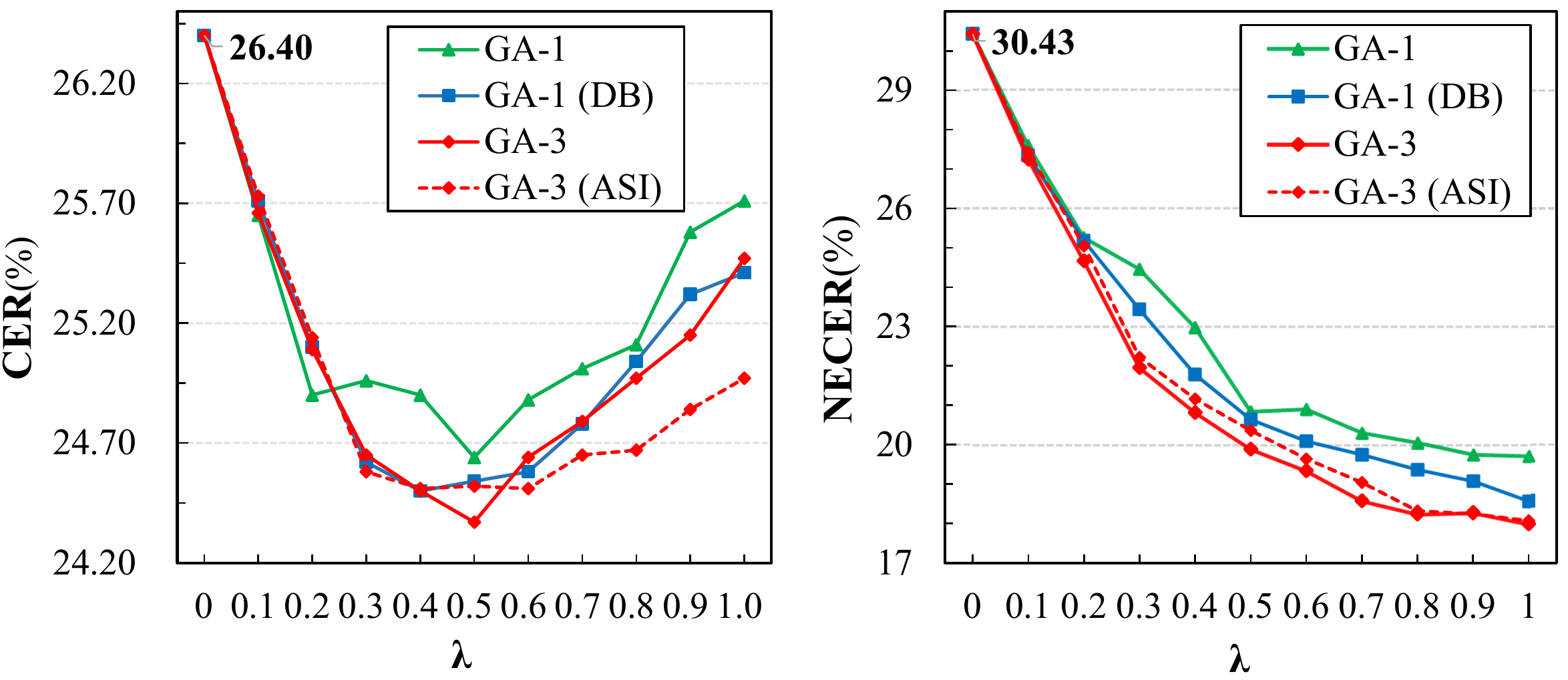}
  \vspace{-10pt}
  \caption{CER and NE-CER with varying $\lambda$ on HKUST-NE (base model): GA-$K$ denotes the gradient accumulation of $K$ c-batches and DB denotes double the size of original batch.}
  \label{Fig.2}
  \vspace{-5pt}
\end{figure}

We first train the base model with the gradient accumulation (GA) from 1, 2, 3, 4 c-batches respectively and conduct evaluations on HKUST-NE. With more c-batches, GA brings better performance. We compare the performance of GA-3 and GA-1 (DB) in Fig.\ref{Fig.2}, because their training costs are close. Fig.\ref{Fig.2} shows that both GA-1 (DB) and GA-3 can achieve lower CER and NE-CER compared with GA-1. Though GA-3 has almost the same trend of CER with GA-1 (DB), it has better NE-CER performance.

On the base model with GA-3, we further investigate the effect of attention scaling interpolation (ASI) on ColDec. As shown in Fig.\ref{Fig.2}, with large $\lambda$, our method suffers from over-biasing that leads to worse CER performance, but ASI effectively relieves over-biasing while introducing limited NE-CER degradation. For the following experiments, we train models with GA-3 and conduct ColDec with ASI. Here, we use $\lambda=0.6$ for all later evaluations on HKUST.

\begin{table}[!ht]
\vspace{-13.5pt}
\centering
\caption{Comparison of the CIF-based model (base and big) before and after applying collaborative decoding (ColDec) on HKUST: CER (\%) and NE-CER (\%)}
\vspace{4.5pt}
\newcommand{\tabincell}[2]{\begin{tabular}{@{}#1@{}}#2\end{tabular}}
\renewcommand\arraystretch{0.83}
\begin{tabular}{lcccc}
\toprule
\multirow{2}{*}{\textbf{Method}} & \multicolumn{2}{c}{\tabincell{c}{\textbf{HKUST-NE}}}  & \multicolumn{2}{c}{\tabincell{c}{\textbf{HKUST-RE}}}\\ 
\cmidrule(r){2-3}\cmidrule(r){4-5}
& \textbf{NE-CER} & \multicolumn{1}{c}{\textbf{CER}}  & \textbf{NE-CER} & \textbf{CER} \\ 
\midrule
\textbf{base} &&&& \\
\quad CIF    & 30.43 & 26.40 & 30.70 & 24.46\\
\quad CIF+ColDec  & 19.64 & 24.51 & 19.56 & 24.38\\
\midrule
\textbf{big} &&&& \\
\quad CIF     & 28.35 & 24.70 & 28.95 & 22.64\\
\quad CIF+ColDec   & \textbf{16.97} & \textbf{22.52} & \textbf{17.57} & \textbf{22.39}\\
\bottomrule
\end{tabular}
\label{tab:Hkust Results 1}
\vspace{-6pt}
\end{table}

As shown in Table \ref{tab:Hkust Results 1}, ColDec brings relative CER reduction of 7.16\% (base) and 8.83\% (big), relative NE-CER reduction of 35.46\% (base) and 40.14\% (big) on HKUST-NE respectively, meanwhile keeps the performance on regular evaluation sets without deterioration. More results for further analysis is given in Table \ref{tab:Hkust Results 2}. The CER (22.92\%) result on HKUST-RE without ASI is worse than that of the result with ASI, which verifies the effectiveness of ASI on suppressing over-biasing on the big model. Then, with ASI removed, we evaluate on HKUST-RE to investigate the performance when only injecting \textit{no-bias}. The result (CER 22.65\%) close to that of CIF baseline without ColDec (CER 22.64\%) shows that \textit{no-bias} effectively models the scenario without contextual injection, and almost keeps original performance. To further explore the improvement of OOV named entity recognition, we then evaluate on HKUST-NE with all OOV NEs and all non-OOV NEs as contextual information respectively. Not surprisingly, the improvement of result with OOV NEs is worse than that of result with non-OOV NEs. Nevertheless, evaluation with OOV NEs still brings 16.74\% NE-CER reduction.

\begin{table}[!ht]
\vspace{-13.5pt}
\centering
\caption{Results for further analysis: NE-CER (\%) and CER (\%) on HKUST (big model). `nb', `oov', `noov' denotes only injecting \textit{no-bias}, injecting all OOV entities (including \textit{no-bias}) and injecting all non-OOV entities (including \textit{no-bias}), respectively. If without additional declaration, all named entities extracted from regular evaluation set are used as the c-batch for evaluation.}
\vspace{4.5pt}
\newcommand{\tabincell}[2]{\begin{tabular}{@{}#1@{}}#2\end{tabular}}
\renewcommand\arraystretch{0.70}
\begin{tabular}{p{1.55cm}p{0.6cm}<{\centering}p{1.215cm}<{\centering}p{0.70cm}<{\centering}p{1.215cm}<{\centering}p{0.70cm}<{\centering}}
\toprule
\multirow{2}{*}{\textbf{Method}} & \multirow{2}{*}{\textbf{ASI}} &\multicolumn{2}{c}{\tabincell{c}{\textbf{HKUST-NE}}}  & \multicolumn{2}{c}{\tabincell{c}{\textbf{HKUST-RE}}}\\ 
\cmidrule(r){3-4}\cmidrule(r){5-6}
\quad & & \textbf{NE-CER} & \multicolumn{1}{c}{\textbf{CER}}  & \textbf{NE-CER} & \textbf{CER} \\ 
\midrule
CIF             & --    & 28.35 & 24.70 & 28.95  & 22.64   \\
\quad + noov    & --    & 20.92 & 24.70 & --      & --     \\
\quad + oov     & --    & 58.26 & 24.70 & --      & --     \\
\midrule
CIF+ColDec      & \Checkmark    & 16.97 & 22.52 & 17.57 & 22.39\\
\quad           & \XSolidBrush      & --     & --     & 16.25 & 22.92\\
\quad + nb      & \XSolidBrush      & --     & --     & --     & 22.65\\
\quad + noov    & \Checkmark    & 10.85 & 23.21 & --     & -- \\
\quad + oov     & \Checkmark    & 41.52 & 23.55 & --     & -- \\
\bottomrule
\end{tabular}
\label{tab:Hkust Results 2}
\vspace{-6pt}
\end{table}

\subsection{Results on AISHELL-2}
\label{ssec:results on aishell-2}

In this section, we validate the effectiveness of our method on a larger dataset (1000h), and analyze two typical cases. On AISHELL-2, we experiment on big model and evaluate with interpolation weight $0.7$ tuned on development set. All results are given in Table \ref{tab:AISHELL-2 Results}. Our method improves the recognition of NEs, from 10.33\% to 5.01\%, with over 50\% relative reduction on AS2-ios-NE, which reveals the potential of our method on larger speech datasets. The results on AISHELL-2 implies that compared with the spontaneous telephone speech in HKUST, the clear speech in AISHELL-2 is helpful in extracting more accurate acoustic and contextual embeddings to achieve better performance.

\begin{figure}[t]
  \centering
  \includegraphics[scale=0.55]{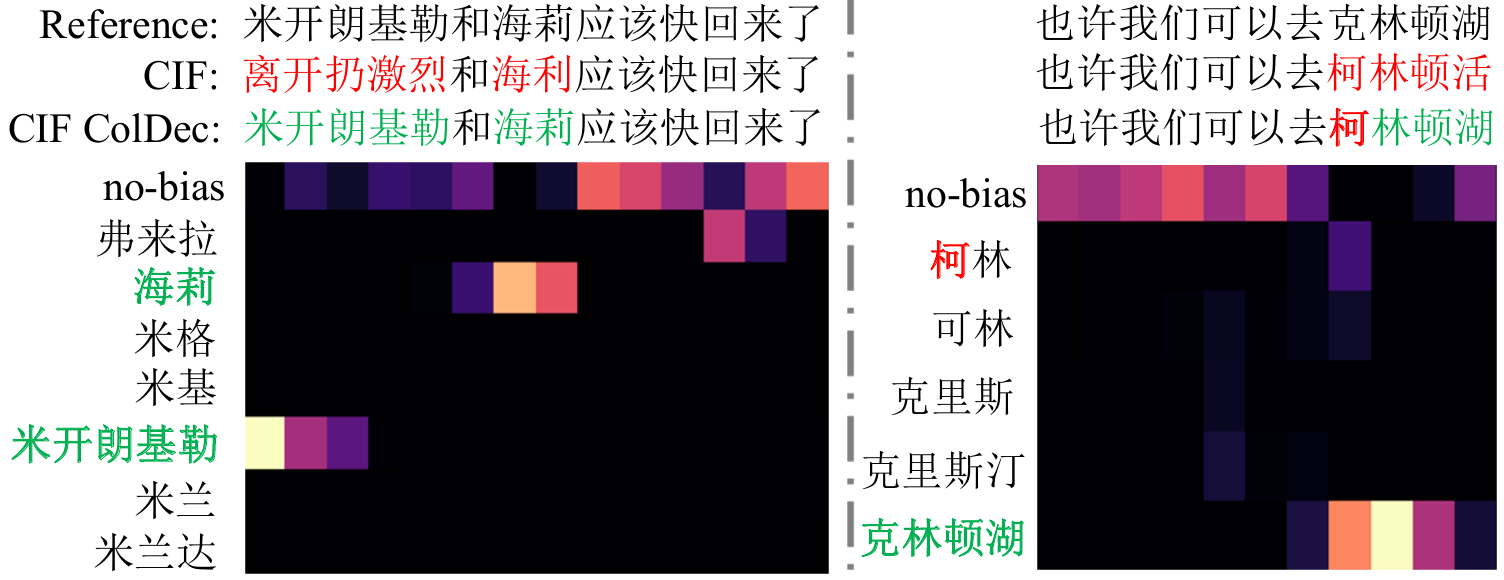}
  \vspace{4.5pt}
  \caption{Typical cases after applying ColDec: The reference, the hypothesis of the CIF-based model before and after applying ColDec are displayed on the top of attention map, while some similar contextual entities are listed on the left of attention map. Larger weights have brighter color, and smaller weights have darker color.}
  \label{Fig.3}
  \vspace{-13.5pt}
\end{figure}

The partial attention maps of two cases are plotted in Fig.\ref{Fig.3}. As depicted in Fig.\ref{Fig.3}, our model pays more attention to entities that include the acoustically relevant character at each step, and attend to \textit{no-bias} option with high confidence when no acoustically relevant information is detected. This proves that our method captures the correlation between acoustic embeddings and contextual embeddings, and leverages this correlation to make biased predictions.

\begin{table}[!ht]
\vspace{-13.5pt}
\centering
\caption{Results on AISHELL-2: NE-CER (\%) and CER (\%).}
\vspace{4.5pt}
\newcommand{\tabincell}[2]{\begin{tabular}{@{}#1@{}}#2\end{tabular}}
\renewcommand\arraystretch{0.83}
\begin{tabular}{lcccc}
\toprule
\multirow{2}{*}{\textbf{Method}} & \multicolumn{2}{c}{\tabincell{c}{\textbf{AS2-ios-NE}}} & \multicolumn{2}{c}{\tabincell{c}{\textbf{AS2-ios-RE}}}\\
\cmidrule(r){2-3}\cmidrule(r){4-5}
& \textbf{NE-CER} & \multicolumn{1}{c}{\textbf{CER}} & \textbf{NE-CER} & \textbf{CER}\\ 
\midrule
CIF         & 10.33             & 6.72              & 10.10             & 5.84 \\
CIF+ColDec  & \textbf{5.01}     & \textbf{5.30}     & \textbf{4.94}     & \textbf{5.70} \\
\bottomrule
\end{tabular}
\label{tab:AISHELL-2 Results}
\vspace{-6pt}
\end{table}

After analyzing hundreds of cases, we find that homophones confuse our model and prevent it from attending to entities accurately. A typical example of homophone confusion is shown on the right side of Fig.\ref{Fig.3}. Since the first two characters of the target NE (highlighted in bold font at the bottom) has similar pronunciation with the first NE along the left column, model also pays notable attention to the counterpart of target NE, and finally emits the first character of its counterpart incorrectly. Besides, we count the frequency of the target NE and its counterpart in training set. We find that the frequency of the target NE is about 1000 times smaller than that of its counterpart, and this fact implies that strong LM learned by the original CIF-based model may also hinder the contextual biasing. In a word, we conjecture that both contextual confusion caused by homophones and strong LM learned by the original model contribute to some failures in our method.

\section{CONCLUSION}
\label{sec:conclusion}

In this work, we propose a collaborative decoding method on the CIF-based model for contextual biasing. Leveraging the acoustic embeddings extracted by the CIF, our method could effectively extract the acoustically relevant contextual information at each decoding step, and then uses a interpolation weight to balance the effect of contextual biasing, thus forming a more controllable approach. Evaluated on two Mandarin ASR datasets which cover 167 and 1000 hours of training data respectively, our method effectively improves the recognition on the contextual entities, and achieves better CER and NE-CER. In the future, we will try to perform more diverse collaborative decodings by using different data sources on the CIF-based model, and to further boost the performance on OOV entities.

\vfill
\pagebreak

\bibliographystyle{IEEEbib}
\bibliography{strings,refs}

\end{document}